# Demographic bias and likelihood-ratio calibration in forensic face comparison, measured with synthetic faces and checked against real ones

Zeno J. M. H. Geradts (1, 2)

(1) Netherlands Forensic Institute, Division of Digital and Biometric Traces, The Hague, the Netherlands. (2) University of Amsterdam, Informatics Institute, Amsterdam, the Netherlands.

Corresponding author: Z. Geradts, z.j.m.h.geradts@uva.nl

**Highlights**

- Synthetic faces from 18 ancestry groups tested eight face recognisers for LR bias
- One LR calibration overstates South Asian and African, understates European faces
- Real faces with identities (BFW, RFW) confirm the direction of the bias
- The size of the bias depends on the image generator and on image quality
- Cohort normalisation narrows but does not remove group differences in LRs

## Abstract

Face recognition scores in forensic casework are increasingly reported as likelihood ratios (LRs). Whether these are equally well calibrated for every demographic group is hard to test, for lack of casework-like data with known identities. I generated 4,991 synthetic persons, balanced over 18 ancestry groups, gender and age, and rendered them as passport photos, edited scenes, degraded CCTV stills and simulated video (689,484 comparisons). A commercial system used in forensic practice and seven open-source recognisers were tested, with LRs calibrated by logistic regression with ELUB bounds, globally and per group. Results were checked on real faces (FairFace, BFW, RFW, SCface) and with a second image generator. On passport photos the groups hardly differed. On CCTV and video all systems ranked the groups in the same way: South and Southeast Asian and Afro-Caribbean faces had two to three times the average false match rate, European faces almost none. One global calibration therefore overstated the LR for the first groups and understated it for European faces, about fivefold between the extremes. Newer recognisers were more accurate, but not fairer. Real faces with identities showed the same direction, with a similar size on BFW and a smaller one on the cleaner RFW images, and a second generator halved the effect. Label-free cohort normalisation reduced but did not remove the differences. Synthetic faces show the direction of demographic bias and are useful for testing mitigation, but do not give its exact size; validation per group and image condition on real material remains necessary.

# 1. Introduction

Images from CCTV systems, doorbell cameras, body-worn cameras and mobile phones have become an everyday source of evidence [40, 41]. In more than thirty years of facial image comparison casework at the Netherlands Forensic Institute I have seen the question from the court change. A verbal opinion based on a morphological comparison is no longer enough; the court, and certainly the defence, want to know how strong the evidence is [39] and how well the method has been validated [22, 23]. Score-based systems whose output can be converted into a likelihood ratio (LR) are one answer to that [10, 11, 31–33]. On good images automated recognisers already perform better than most human examiners [21], and laboratories use them more and more in CCTV and video cases. The value of such an LR stands or falls with its calibration. An LR of 1,000 should mean that the observed score is 1,000 times more probable if the two images show the same person than if they show different persons, and that should hold for the case at hand, not just on average [8, 9].

It is well known that face recognisers do not perform equally for all demographic groups. In the NIST FRVT demographic study false match rates differed by one to two orders of magnitude between groups, with the highest rates for West and East African and East and South Asian faces in many algorithms [1], and similar effects have been reported on research benchmarks such as RFW [7] and in other evaluations and surveys [34, 35]. For forensic use this has a consequence beyond a different error rate. If one calibration is trained on a mixed population, the LR will be too high for groups whose different-source scores are high, and too low for the others. Such a shift does not show up in an overall Cllr.

Testing this under forensic conditions is not easy. Databases with many persons per demographic group, recorded under CCTV and video conditions, hardly exist. The sets that are available are small, dominated by one population or recorded at good quality; SCface, for example, has 130 mostly Croatian subjects [6]. Collecting new data is expensive, and under the GDPR biometric data used to identify a person are a special category that needs a specific legal basis [24]. The EU AI Act classifies remote biometric identification and AI used to evaluate evidence in criminal investigations as high-risk, and requires that training, validation and test data are examined for possible bias and are as representative as possible (Art. 10) [24]. Laboratories that use these systems therefore need a way to show how their system and its LRs behave for each group and image condition. In court the question is concrete: does the method work as well for this suspect, who may belong to a group that is underrepresented in the validation data? Until now that question could only be answered in general terms.

Synthetic faces could fill this gap. They do not belong to existing persons, can be generated in any number for any group, can be rendered under controlled conditions and can be shared without privacy restrictions [14, 15, 38]. The difficulty is that a bias measured on synthetic faces may come from the recogniser, but also from the generator. Diffusion models inherit the biases of their training data and may make some groups more uniform than others, which by itself raises the false match rate within those groups.

In this study I address four questions. Do a forensic system and recent open-source recognisers differ between demographic groups in false match rate, false non-match rate and LR calibration when they are tested on synthetic faces in passport, CCTV and video conditions? How much of the effect is caused by the generator, judged against real faces (FairFace, BFW, RFW, SCface) and a second generator? Can a score normalisation that uses no demographic labels reduce the differences? And can an LR calibration built on synthetic data be used on real surveillance images? At the end I discuss the use of synthetic faces for proficiency testing.

# 2. Materials and methods

## 2.1 Synthetic identities

I defined 18 ancestry groups, among them Northern European, Southern European, West African, East African, Afro-Caribbean, South Asian, Southeast Asian, Indonesian, East Asian, Middle Eastern and Latin American. These were crossed with gender presentation and five age bands, and for each person the other attributes were drawn at random: hair, facial hair, glasses, build and skin tone on the Monk scale [28]. For each description a reference passport photo was generated with Z-Image-Turbo or Qwen-Image-2512 [26, 27]. A vision-language model then checked whether the photo matched the description in ancestry, age, gender and skin tone (with a tolerance of one Monk step), and photos that did not were rejected. Persons that were too similar to another person (ArcFace R50 cosine of 0.5 or more between the references) were removed. This left **4,991 persons**.

## 2.2 Conditions

Each person was placed by Qwen-Image-Edit-2511 (Lightning) in 13 everyday situations, such as a street, a shop, public transport, indoors and at night. Every edit was kept at high quality (*edit_hq*) and was also degraded into two kinds of CCTV still. In *cctv_mid* the face is about 64 pixels wide, with blur, JPEG quality 55 and noise; in *cctv_low* it is about 32 pixels wide with JPEG quality 35. In addition, short clips were simulated for six camera types (entrance, ceiling, distant, backlit, body-worn and phone) with H.264 compression, from which I used both I- and P-frames. A second passport photo (*ref2*) gave a mated passport pair. The set contains 215,341 images and 689,484 pairs. Mated pairs compare the reference with all images of the same person, non-mated pairs compare it with images of up to 40 persons of the same ancestry and gender.

## 2.3 Recognisers

Eight systems were tested. The first is a commercial face recognition system used in forensic practice, referred to as System C, with its minimum eye distance set to 0.015. The others are ArcFace R50 (buffalo_l) [2], which was also used to build the set so that its results are optimistic, AdaFace IR-101 WebFace12M and IR-50 MS1MV2 [3], AdaFace ViT-B WebFace4M, ArcFace IR-101 WebFace4M [16], AdaFace ViT-B KP-RPE WebFace12M [4] and Glint360K R100 (antelopev2) [17]. For the open systems faces were detected with SCRFD [20], or with YuNet [19] where SCRFD failed, aligned to the usual 112 × 112 template, and compared by cosine similarity. System C used its own face finder. Image quality was estimated with eDifFIQA [18].

## 2.4 Error rates

For each condition class (passport, edit_hq, cctv_still and video) one threshold was set at an FMR of $10^{-3}$ over all non-mated pairs. FMR and FNMR were then computed per group, with 95% confidence intervals from a bootstrap over persons (1,000 resamples). The failure-to-enrol rate (FTE) was also recorded per group.

## 2.5 Likelihood ratios

Scores were converted to LRs by logistic regression with ELUB bounds [9], as in earlier work on face comparison [31, 32], in five-fold cross-validation in which no person appeared in both the training and the test part. This was done once for all groups together and once for each ancestry group separately. I call the mean difference in log10 LR between the global and the group-specific calibration over the mated pairs of a group, taken relative to the mean over all groups, the **LR shift** of that group. The **spread** is the difference between the highest and the lowest group shift. Performance was summarised with Cllr and Cllr_min [8], overall and per group.

## 2.6 Confounders

To see whether the group differences are really about ancestry and not about image quality or appearance, I fitted a mixed model of the non-mated score with ancestry, gender, age band, quality, glasses, headwear and facial hair as fixed effects and person as random effect. A quality-aware calibration was also tested, in which the quality of both images was added to the logistic regression.

## 2.7 Score normalisation

Z-norm, T-norm and S-norm [12] and a top-*k* T-norm [13] were tested. In top-100 T-norm each score of a probe is normalised with the mean and standard deviation of the 100 highest scores of that probe against a cohort of 600 faces of mixed origin. No demographic labels are needed for this.

## 2.8 Checks of the generator

*FairFace.* From the FairFace training set I used the adult faces (65,351 images in seven race categories) [5]. FairFace has no identity labels, so I could only look at false matches. A gallery of 357 images per race and gender was compared with all faces of the same race and gender as probes, leaving out the image itself and near-duplicates (cosine 0.75 or more on the original images). The threshold was set where the mean FMR over the groups is $10^{-3}$. The test was repeated with the faces degraded in the same way as the CCTV stills, and with T-norm. System C was run on the original FairFace images on a separate computer, and only group averages were kept. To compare the uniformity of synthetic and real groups, I expressed the mean non-mated similarity of each synthetic group relative to the corresponding FairFace group.

*Second generator.* The same 1,008 descriptions were also generated with FLUX.2 [klein] 4B (Apache 2.0) as passport photos, edits and CCTV stills, and analysed with the same pair design, both for all persons and for the 593 persons left after removing near-duplicates.

*Real faces with identities.* Two sets with real identities and group labels were used. BFW [30] contains 800 persons with 25 images each, balanced over Asian, Black, Indian and White, female and male. From the RFW test set [7] I drew 1,000 persons at random for each of its four groups (14,305 images). The first image of each person served as reference and the other images, at original quality and degraded like the synthetic CCTV stills, as questioned images. Non-mated pairs were formed within the same group, with all 99 other persons of the subgroup for BFW and 20 others for RFW. This gave 19,200 mated and 1.9 million non-mated pairs for BFW, and 10,306 and 206,479 for RFW. Scores were used as they were and after T-norm and top-100 T-norm with the cohort of 600 synthetic references. LRs, LR shift, spread and Cllr were computed in the same way as for the synthetic set. AdaFace IR-50 MS1MV2 was left out on RFW, because RFW persons also occur in MS-Celeb-1M on which that model was trained. I also tried LFW, with groups predicted by a classifier trained on FairFace, but with 54% accuracy that classifier was too poor to be of use.

*SCface.* SCface [6] contains 130 persons photographed with a mugshot camera and with surveillance cameras at 1.0, 2.6 and 4.2 m (198,237 pairs, 1,525 of them mated). The System C scores were T-normalised with the cohort of 600 synthetic references and then calibrated with the calibration trained on the synthetic CCTV stills. I compared this with a calibration trained on SCface itself, again with cross-validation over persons. Other combinations of normalisation and calibration are described in an internal validation report.

*Memorisation.* Each synthetic reference was compared with 69,719 real faces, from FairFace and other sources, to check that the generator had not reproduced a real person.

### 2.9 Software and data

The analyses were done in Python (NumPy, pandas, scikit-learn, onnxruntime and PyTorch), with the NFI *lir* package for calibration [29]. The code is available at https://github.com/geradts1/facebias under the MIT licence. The synthetic image set will be released with the journal version. FairFace, BFW, RFW and SCface can be obtained from their owners.

## 3. Results

### 3.1 Face finding

With SCRFD alone no face was found in 14.9% of the images. Using YuNet as a second detector brought this down to 3.4%. System C with its default minimum eye distance of 0.10 failed on 22% of the images, mostly the small faces in CCTV stills and video, and at 0.015 this dropped to 3.3%. A setting of the face finder can thus quietly remove most of the difficult images from an evaluation.

### 3.2 False match and false non-match rates

On passport photos the differences between groups were negligible for all systems. On CCTV stills and video they were large, and they pointed the same way in every system. South and Southeast Asian, Indonesian and Afro-Caribbean faces had two to three times the average FMR, while for the European groups it was close to zero. For System C on video the FMR relative to the average was 2.5 (95% CI 1.4–3.7) for Indonesian and 2.0 (1.1–3.1) for Afro-Caribbean faces. Women had 1.4 times the FMR of men, and persons aged 60–74 twice the average. The FNMR of System C at an FMR of $10^{-3}$ was 13.2% on edits, 17.1% on CCTV stills and 44.0% on video (Table 2).

The systems also agreed on the order of the groups. Spearman correlations between systems were 0.83–0.88 on video and 0.64–0.74 on edits and CCTV stills, and 15 of the 18 groups deviated in the same direction in all seven systems with complete results. The newest recogniser, AdaFace KP-RPE, had the best Cllr on edits and CCTV stills, yet the largest LR spread on video (0.81) and an FNMR of 44.5% there. Higher accuracy did not come with smaller differences between groups.

## 3.3 Likelihood ratio calibration

With one global calibration the LR was too high for the groups with high non-mated scores and too low for the others. Averaged over seven systems the relative LR shift ranged from +0.33 log10 for Indonesian to −0.41 for Northern European faces. That is a spread of about 0.75 log10, or more than a factor five between the extremes. For System C the spread was 0.65 on video and 0.87 on CCTV stills. Table 1 gives Cllr per system and condition. System C had the best Cllr on video (0.347) and KP-RPE on edits (0.090) and CCTV stills (0.115). On CCTV stills Cllr differed by a factor 1.8–2.4 between the worst and the best group (System C: 0.08–0.19). Calibrating per group did not improve the overall Cllr.

## 3.4 Quality and confounders

A calibration that took image quality into account lowered the Cllr of AdaFace on video from 0.44 to 0.26, but the LR spread stayed the same. In the mixed model the ancestry effect on non-mated scores went down only from 0.42 to 0.35 standard deviations after correction for quality, gender, age and attributes. Image quality does not explain the differences between the groups.

## 3.5 Effects of the generator

*Homogeneity.* The generator did not make persons equally distinct in every group. With Z-Image-Turbo 70% of the West African candidates were removed as near-duplicates, against 26% of the Northern European ones. Over the groups, this homogeneity correlated with the FMR ($\rho = 0.71$).

*FairFace.* The order of the groups in non-mated similarity was the same for synthetic and real faces (rank correlation 0.71–0.96 over the systems). Compared with their real counterparts, however, the synthetic non-white groups were about twice as over-uniform as the white groups.

*Second generator.* FLUX.2 [klein] made the faces much more alike than Qwen did (mean non-mated similarity 0.33 against 0.15, with 40% near-duplicates), but the differences between groups were smaller (homogeneity gap 0.08 against 0.13; excess uniformity of non-white relative to white groups 1.1 against 2.2). Its group order agreed less with real faces ($\rho$ 0.36–0.46 against 0.75–0.82 for Qwen), and its LR spread was about half as large: 0.37–0.48 against 0.70–0.94 for all persons, and 0.52–0.65 against 0.83–1.00 for the 593 persons kept in both sets. The size of the bias that is measured depends on the generator, but its direction does not.

*Look-alikes and twins.* Synthetic twins, generated from the same description with another seed, were matched in 47–66% of the passport comparisons and received an LR above 1 in 73–82%. For look-alike pairs this was 74–84%, and 7–9% even received an LR above 100.

### 3.6 Real faces

On FairFace, System C showed the same order of groups as on the synthetic set. The FMR relative to the median was 2.5 for Indian, 2.1 for Southeast Asian, 1.2 for East Asian, 1.0 for Black and Middle Eastern, 0.7 for Latino and 0.3 for White faces, a ratio of 8.7 between highest and lowest (15.7 with a duplicate cut-off of 0.80). Women had 2.5 times the FMR of men, and persons of 50 and over 2.6 times the average. The rank correlation between synthetic and real group FMR was 0.75 (95% CI 0.36–0.93) on edits and 0.57 on CCTV stills. For the seven open systems the ratio between the highest and lowest group was 2.8–3.6 on photos and 3.2–3.9 at CCTV quality (Table 3).

*Mated pairs on real faces (BFW, RFW).* Both sets confirm the direction found with synthetic faces (Table 5). With one global calibration the LRs of Indian and African/Black faces were too high and those of White/Caucasian faces too low, and White faces had by far the lowest FMR (0.13–0.17 times the average). On BFW the size of the effect was close to that on the synthetic set. The LR spread was 0.77–1.17 log10 over six recognisers, against about 0.75 on synthetic faces, and White faces had about half the Cllr of the other groups (0.11 against 0.20–0.26). Women had 1.7–2.7 times the FMR of men. On RFW the FMR differences were even larger (6–19 between highest and lowest group on the originals, 9–20 at CCTV quality), but the LR spread was small (0.21–0.36). On these good-quality images mated and non-mated scores lie far apart in every group, which leaves little room for a shift. Asian faces were overstated on BFW (+0.16) and slightly understated on RFW (−0.12). Degrading the images to CCTV quality changed little on BFW, whose images are already small crops. T-norm with the synthetic cohort halved the FMR ratio on RFW (to 4–9) and reduced the LR spread on BFW by 20–30%, but did not remove it. On BFW top-100 T-norm slightly increased the FMR ratio between women and men (to 2.1–2.7).

**Table 5.** Real faces with identities: mean over six open recognisers, original quality, one global calibration. FMR relative to the average at FMR $10^{-3}$; LR shift = mean log10 LR of the global minus a per-group calibration (mated pairs), relative to the group mean.

| Group | BFW FMR ratio | BFW LR shift | BFW Cllr | RFW FMR ratio | RFW LR shift | RFW Cllr |
|---|---|---|---|---|---|---|
| Indian | 1.70 | +0.38 | 0.207 | 1.14 | +0.12 | 0.045 |
| Asian | 1.41 | +0.16 | 0.258 | 1.07 | −0.12 | 0.050 |
| Black / African | 0.72 | 0.00 | 0.202 | 1.63 | +0.10 | 0.027 |
| White / Caucasian | 0.17 | −0.54 | 0.114 | 0.13 | −0.11 | 0.033 |
| LR spread (range over systems) | | 0.77–1.17 | | | 0.21–0.36 | |

### 3.7 Mitigation

On the synthetic video, top-100 T-norm reduced the LR spread from 0.75 to 0.37, the ratio between the highest and lowest group FMR from 33 to 12, and the FNMR from 53% to 30%. On CCTV stills the spread went from 0.69 to 0.56 and on edits from 0.59 to 0.47. The difference between women and men did not become smaller. On real faces at CCTV quality, T-norm reduced the ratio between highest and lowest group FMR from 3.2–3.9 to 1.8–2.8 with the whole cohort, and to 1.8–2.3 with top-100. Southeast Asian faces went from 1.7–2.4 to 0.8–1.1 times the median, and the ratio between women and men from 1.4–1.7 to 1.1–1.3.

### 3.8 Transfer to real surveillance images

On SCface, the calibration trained on synthetic CCTV stills, used after T-norm with the synthetic cohort, gave a Cllr of 0.020 (95% CI 0.012–0.032). A calibration trained on SCface itself gave 0.018 (0.010–0.028), and at 4.2 m the two gave 0.075 and 0.070. Misleading evidence occurred in 0.5% of the pairs in both directions, for both calibrations. Without normalisation the transferred calibration was conservative (Cllr 0.13), while top-100 T-norm made it unsafe, with 4–7% of the non-mated pairs receiving an LR above 1. This combination was chosen after several had been compared, so it needs to be confirmed on other real CCTV sets.

### 3.9 Controls

The highest similarity of any synthetic reference to the 69,719 real faces was 0.835 (ArcFace), below the level of mated passport pairs. The similarity of each synthetic person to its nearest neighbour (0.85–0.93) was much higher than among real persons (0.25–0.37), which is another sign of the over-uniformity described above.

## Tables

**Table 1.** Cllr per recogniser and condition class (global logistic calibration with ELUB, identity-disjoint cross-validation; synthetic set). Ordered by Cllr on CCTV stills.

| Recogniser | Edits (HQ) | CCTV stills | Video |
|---|---|---|---|
| AdaFace ViT-B KP-RPE (WebFace12M) | 0.090 | 0.115 | 0.362 |
| System C (commercial, forensic) | 0.115 | 0.127 | 0.347 |
| ArcFace R50 (buffalo_l)[a] | 0.114 | 0.135 | 0.410 |
| AdaFace ViT-B (WebFace4M) | 0.125 | 0.144 | 0.406 |

| Recogniser | Edits (HQ) | CCTV stills | Video |
|---|---|---|---|
| Glint360K R100 | 0.131 | 0.156 | 0.438 |
| AdaFace IR-101 (WebFace12M) | 0.144 | 0.161 | 0.423 |
| ArcFace IR-101 (WebFace4M) | 0.141 | 0.166 | 0.435 |
| AdaFace IR-50 (MS1MV2) | 0.197 | 0.204 | 0.500 |

[a] Used to filter near-duplicates when building the set; results optimistic.

**Table 2.** System C (minimum eye distance 0.015) on the synthetic set: FNMR at FMR $10^{-3}$, LR spread (highest minus lowest relative log10 LR shift over 18 ancestry groups) and highest group FMR relative to average (95% CI).

| Condition | FNMR | LR spread (log10) | Highest group FMR ratio (95% CI) |
|---|---|---|---|
| Edits (HQ) | 13.2% | 0.42 | Afro-Caribbean 3.7 (1.2–6.8) |
| CCTV stills | 17.1% | 0.87 | Afro-Caribbean 3.4 (1.6–5.9) |
| Video | 44.0% | 0.65 | Indonesian 2.5 (1.4–3.7) |

**Table 3.** Real faces (FairFace train, adults): FMR ratios for seven open-source recognisers (range over systems), at the threshold where the mean group FMR is $10^{-3}$.

| | Photo | CCTV low | + T-norm (whole cohort) | + top-100 T-norm |
|---|---|---|---|---|
| Highest vs lowest group | 2.8–3.6 | 3.2–3.9 | 1.8–2.8 | 1.8–2.3 |
| Southeast Asian vs median | 1.6–2.1 | 1.7–2.4 | 1.1–1.7 | 0.8–1.1 |
| Women vs men | 1.5–1.8 | 1.4–1.7 | 1.1–1.4 | 1.1–1.3 |

**Table 4.** Generator comparison on the same 1,008 descriptions (open recognisers, range over systems).

| | Qwen-Image-2512 | FLUX.2 [klein] 4B |
|---|---|---|
| Mean non-mated similarity | 0.15 | 0.33 |
| Near-duplicates removed | – | 40% |

| | Qwen-Image-2512 | FLUX.2 [klein] 4B |
|---|---|---|
| Homogeneity gap between groups | 0.13 | 0.08 |
| Excess uniformity non-white / white | 2.2 | 1.1 |
| Rank correlation with real faces (ρ) | 0.75–0.82 | 0.36–0.46 |
| LR spread, all persons | 0.70–0.94 | 0.37–0.48 |
| LR spread, 593 persons kept in both | 0.83–1.00 | 0.52–0.65 |

*Figures 1–5 below.*

**Fig. 1.** One synthetic person (not used in any test material) in the conditions of the study: passport photos, an edited night-street scene, two degraded CCTV stills and frames from simulated analogue, infrared and body-worn video.

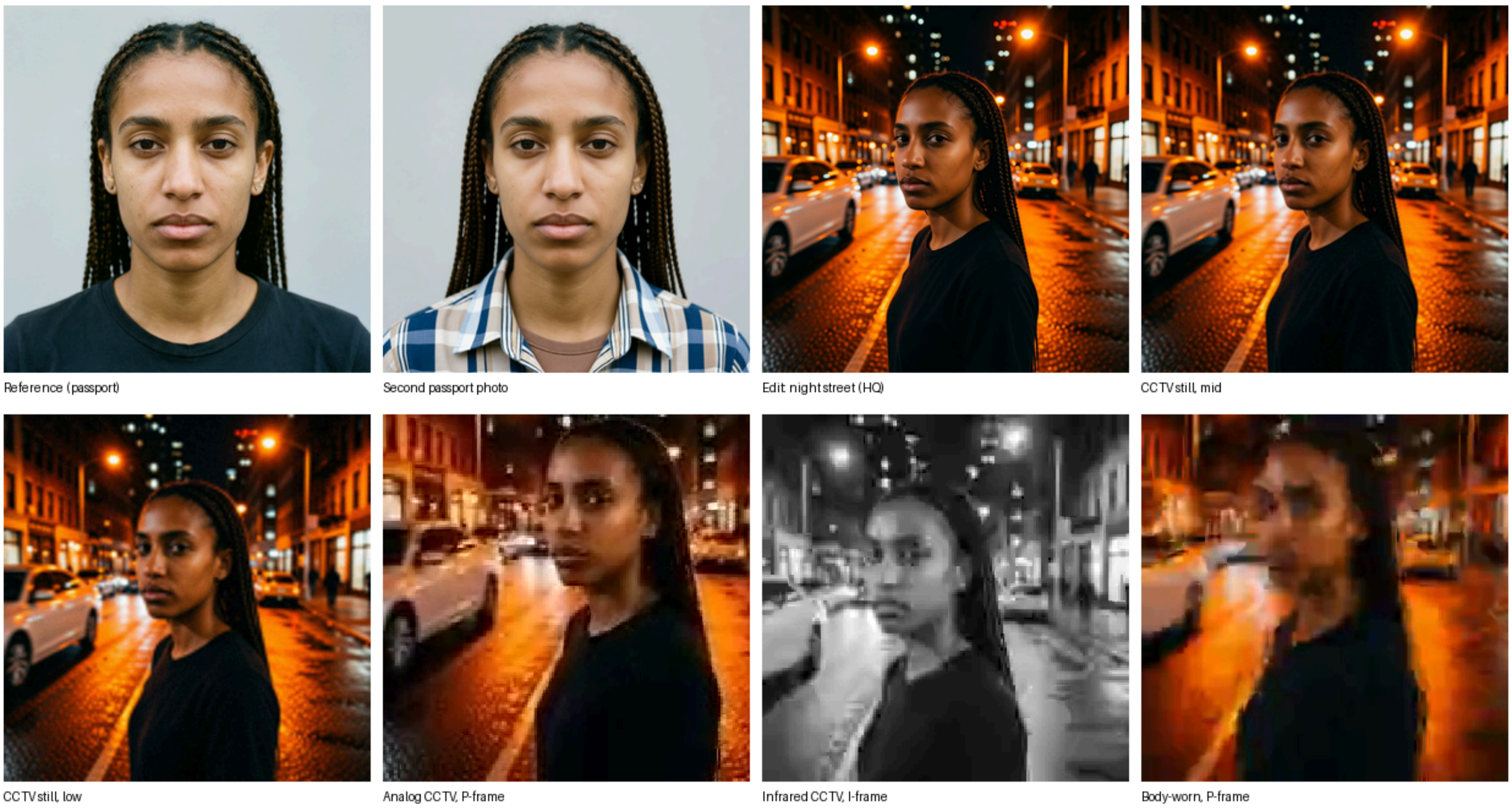

**Fig. 2.** False match rate per ancestry group, System C, video.

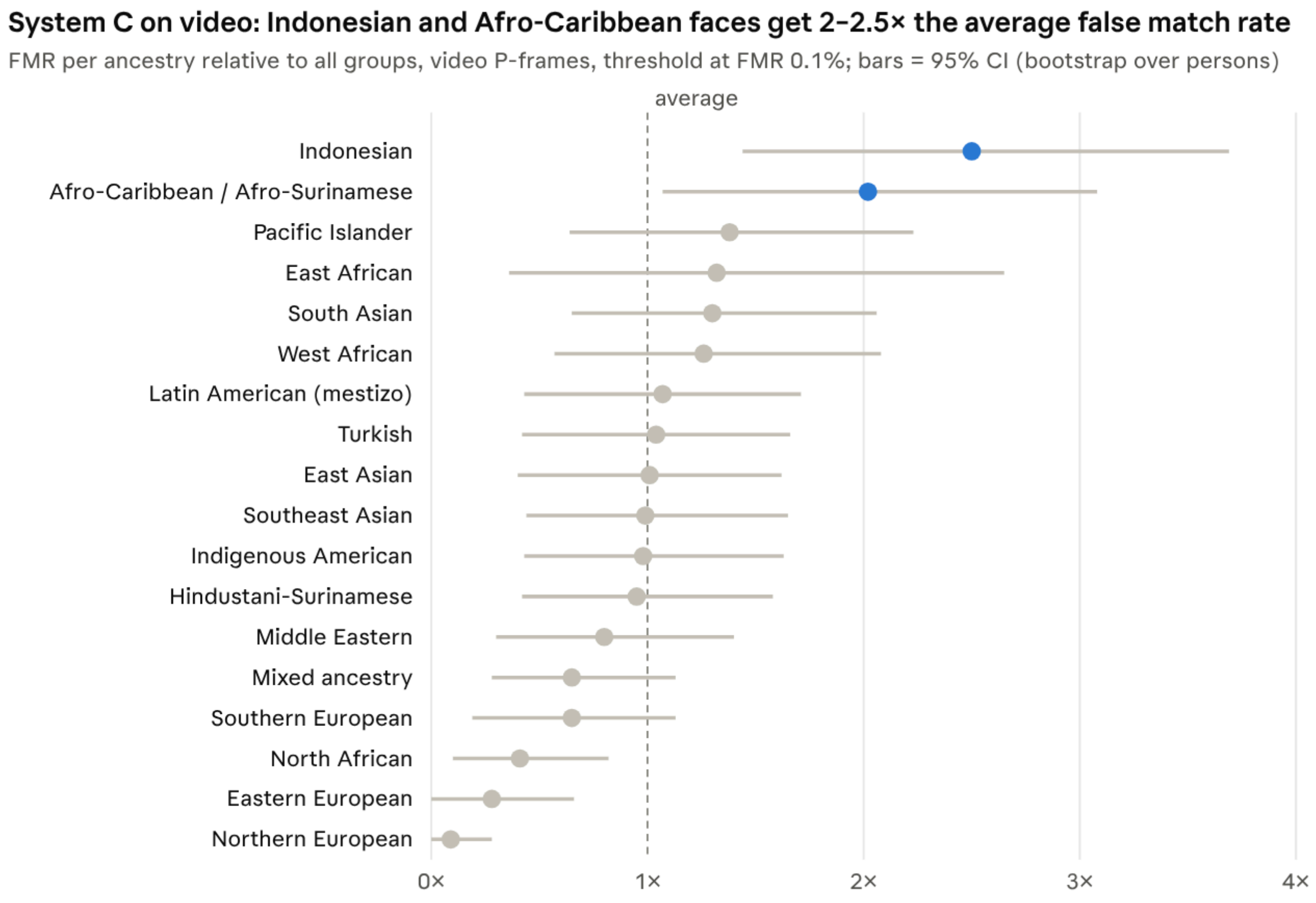

System C, video (P-frames), per ancestry group; 4,991 synthetic persons; Supplementary Table S3

**Fig. 3.** LR shift per ancestry group, System C, CCTV stills.

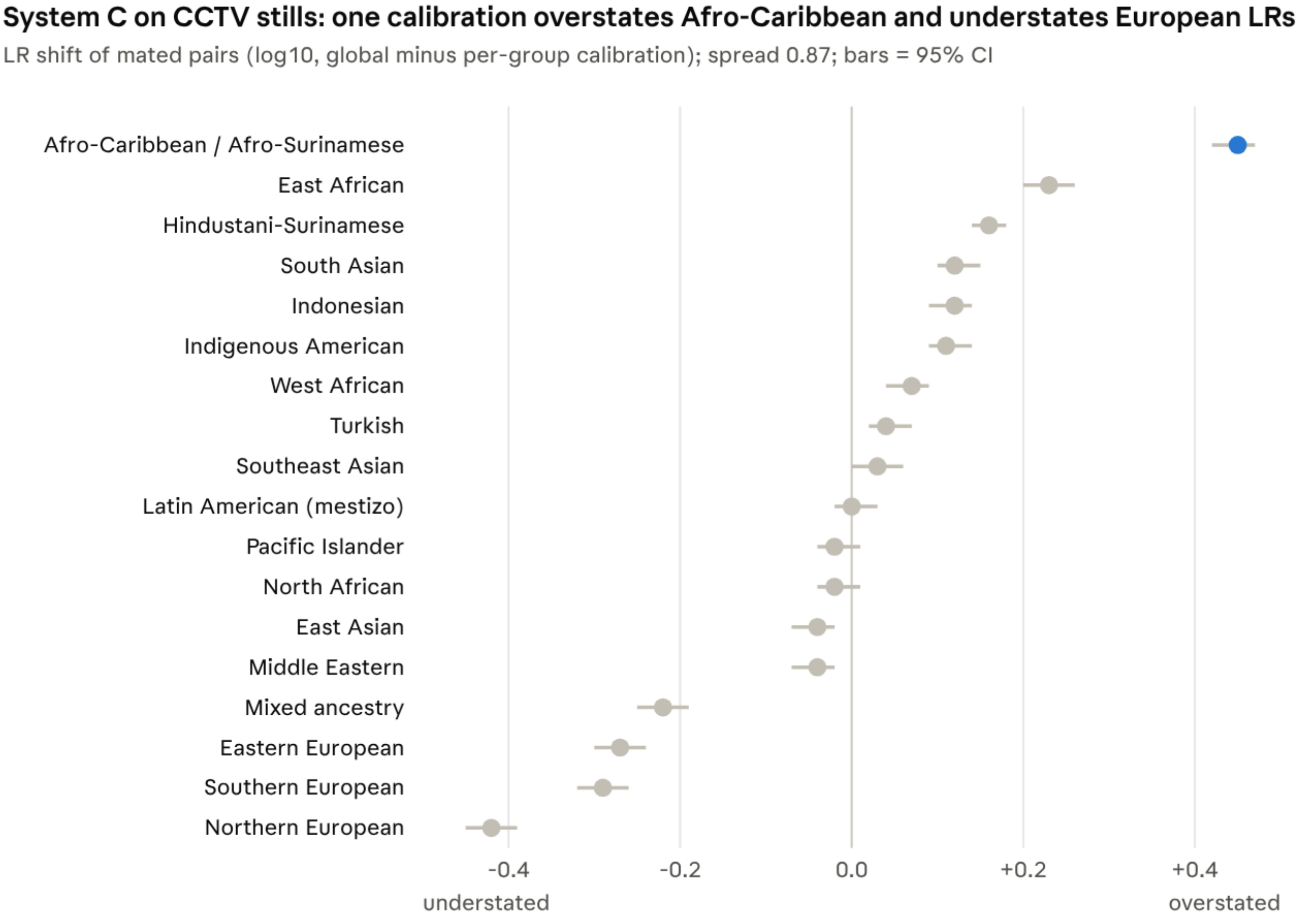

System C, CCTV stills, per ancestry group; identity-disjoint CV, 10 repeats, joint bootstrap; Supplementary Table S4

## **Fig. 4.** LR shift per group on synthetic and real faces.

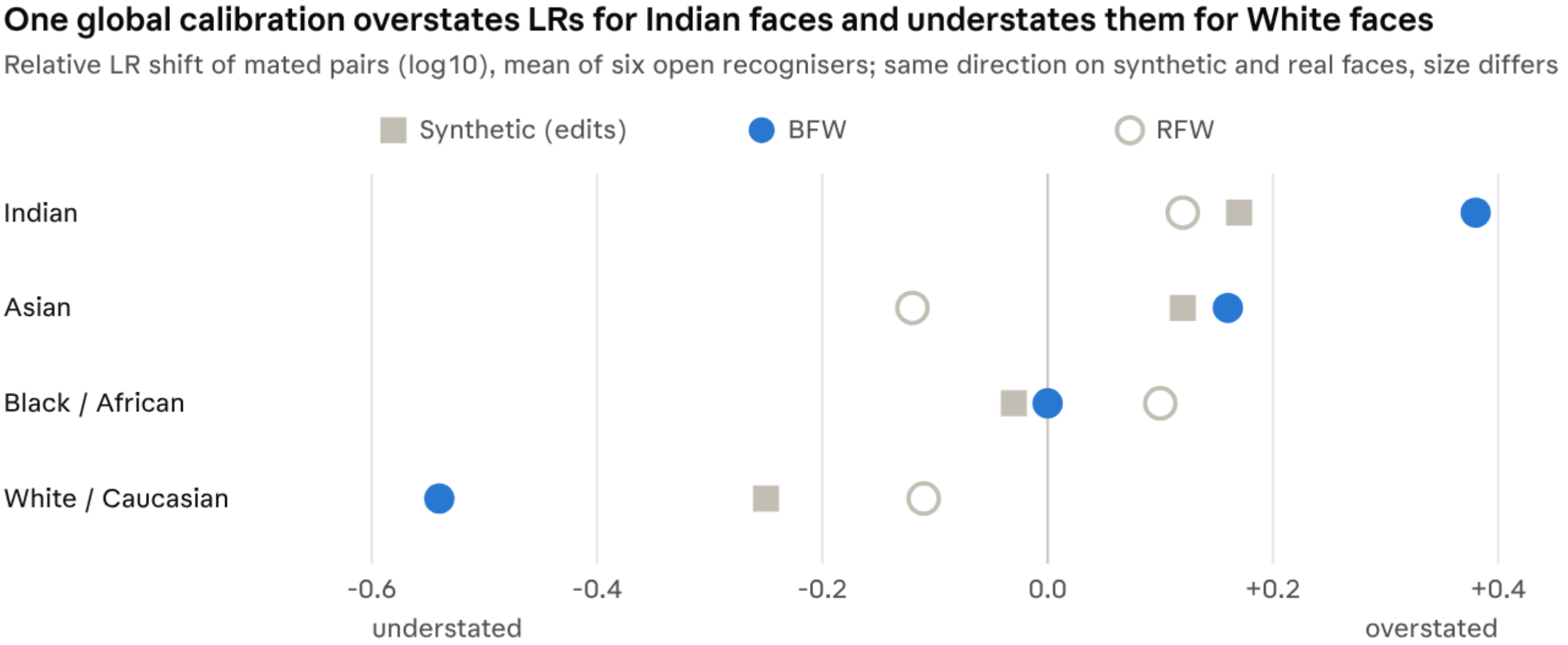

Synthetic edits (11 ancestry groups merged into 4), BFW and RFW at original quality; six open recognisers; Supplementary Tables S4 and S8

**Fig. 5.** Effect of cohort normalisation on the false-match gap between groups on real faces. Thresholds differ by set: FairFace at the mean-group FMR of 0.1%; BFW and RFW at a global FMR of 0.1%.

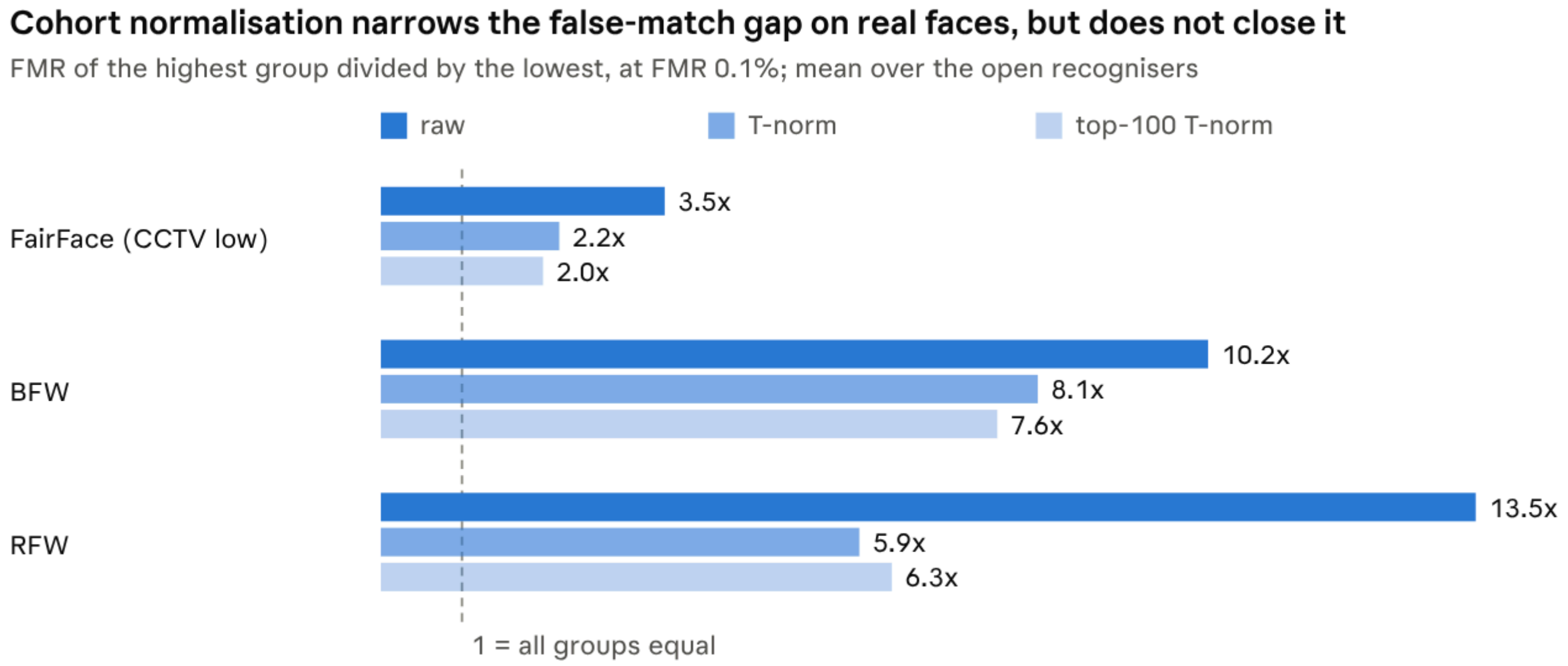

FairFace at CCTV-low quality (7 recognisers); BFW and RFW at original quality (6 recognisers); Supplementary Tables S7 and S8

# 4. Discussion

## 4.1 What synthetic faces can and cannot show

Eight recognisers, including a commercial system used in forensic practice and a transformer from 2024, agreed on which groups receive more false matches in CCTV and video conditions, and System C showed the same order on 65,000 real faces. On BFW and RFW, two real sets with identities, one global calibration overstated the LRs of Indian and African faces and understated those of White faces, just as it did on synthetic faces. I therefore consider the direction of the bias a property of the recognisers and not something the generator introduced. Only the position of Asian faces differed between the two real sets. The size of the bias is another matter, as it depends on the material. On BFW the LR spread (0.8–1.2 log10) was close to the synthetic estimate, while on the good-quality RFW images it was about a third of it. The generator also plays a part. The homogeneity of the generated groups correlated with their FMR, synthetic non-white groups were about twice as over-uniform as white groups when compared with real faces, and a second generator halved the LR spread. Synthetic faces are thus well suited to detect bias, to compare systems and to test mitigation. The absolute FMR ratios and LR shifts they give, however, should not be presented as properties of a system in casework. In my view a synthetic evaluation should always include at least one set of real faces as an anchor, here FairFace, and a second generator.

## 4.2 Consequences for LRs in casework

With one global calibration the LR for a suspect from a group with high non-mated scores is too high, for the most affected groups on average by about a factor two on CCTV and video, and the LR for a European face is too low. This is a systematic error that a pooled Cllr does not show. Calibrating per group is not a practical way out. It requires that the group of the suspect is known and verified, it brings demographic labels into an evidential calculation, and it did not improve Cllr in this study. Reporting Cllr and LR shift per group and condition, and using a normalisation that needs no labels, are easier to defend. Score normalisation and calibration methods aimed at fairness have been proposed for biometric verification [36, 37], but as far as I know they have not yet been evaluated for forensic LRs. Top-100 T-norm halved the spread on synthetic faces and the FMR ratio on real faces at CCTV quality, but on SCface it made the calibrated LRs less safe (4–7% of non-mated pairs with an LR above 1). Plain T-norm with a synthetic cohort, on the other hand, transferred well. A normalisation therefore needs its own validation before it is used for LRs.

The face finder deserves more attention than it usually gets. A default minimum eye distance silently excluded 22% of the images, mainly the small faces from CCTV and video, which are exactly the conditions in which the bias appears. A validation report should state the face finder settings and the FTE per group. In casework this matters directly, because the images on which a system fails are often the very images on which the expert then has to rely on his own comparison.

## 4.3 Newer systems

The most accurate recogniser, KP-RPE, was not the fairest. It had the best Cllr on edits and CCTV stills, but the largest LR spread on video. Laboratories should not assume that an upgrade removes demographic differences, and each new version needs its own validation per group.

## 4.4 Regulatory context

Under the AI Act, face recognition for remote biometric identification and AI used by law enforcement to evaluate the reliability of evidence are high-risk systems (Annex III). Providers have to examine their data for bias (Art. 10) and document accuracy for relevant groups, and deployers in law enforcement have to carry out a fundamental rights impact assessment (Art. 27). The GDPR and the Law Enforcement Directive restrict the processing of biometric data for these purposes. Synthetic faces make such examinations possible without processing data of real persons, on two conditions. No generated face may reproduce a real person, which I checked against 69,719 real faces, and the licence of the generator must allow biometric research; one candidate generator was dropped for that reason. The EDPB has pointed out that models trained on personal data may themselves contain personal data [25]. For that reason I tested the output of the generators for memorisation, and for the real data sets I report only group averages.

## 4.5 Proficiency testing

Synthetic faces can also be used for proficiency tests. Within the ENFSI Digital Imaging Working Group this has been a recurring problem, because good test material with known ground truth is scarce and real casework images cannot be shared. The ground truth is known, the demographic composition can be chosen freely, the images can be degraded like casework material, there are no consent issues, and new items can be made at any time, so that a test cannot be learned. A first pilot with 40 items, selected with four recognisers and with matched attributes for the different-person items, also showed where the problems lie. Generated images may contain readable text, accessories or artefacts that give the answer away, and twins generated from the same description are not real siblings. Items should therefore be selected with several recognisers, checked automatically for text and visually by an expert, and compared with real casework-like image pairs before they are used. A study with examiners is planned and will be reported separately.

## 4.6 Limitations

This study has several limitations. The ancestry labels are the descriptions given to the generator and not a self-identification, while the FairFace labels are judgements of annotators and the BFW and RFW labels were assigned by the creators of those sets. The degradations and the video were simulated, so the motion blur, compression artefacts and pose of real cameras are only partly represented. The synthetic set is over-uniform, and mated variation such as ageing or a real change of expression is limited, which makes the mated scores optimistic. ArcFace R50 was used to build the set, so its results are optimistic as well. FairFace has no identity labels, which means that only the FMR could be studied there. The real mated pairs come from BFW and RFW, which consist of web images of celebrities with annotator labels and only four groups, and from SCface, which is small, mostly of one ancestry and too small for an analysis per group. System C was not run on degraded FairFace images or on BFW and RFW. Finally, the results apply to the versions of the systems that were tested.

# 5. Conclusions

On CCTV and video, all eight recognisers in this study give more false matches for South and Southeast Asian and Afro-Caribbean faces and fewer for European faces, while on passport photos the differences are negligible. A single global LR calibration then overstates the evidence for the first groups and understates it for European faces, and a pooled Cllr hides this, so Cllr and LR shift should be reported per group and image condition.

Synthetic faces reliably show the direction of this bias, as FairFace, BFW and RFW confirm. Its size depends on the generator and on image quality: it matched on BFW and was smaller on the clean RFW images. The generator should therefore be part of the validation. Top-100 T-norm, which needs no demographic labels, roughly halves the differences between groups on synthetic and on real faces at CCTV quality, but LRs computed after normalisation need their own validation. A calibration built on synthetic CCTV stills, combined with plain T-norm, worked on real surveillance images (SCface) almost as well as a calibration trained on those images. Face finder settings have a large effect on which images end up in an evaluation and should always be reported. Finally, synthetic faces look promising as material for proficiency tests, provided that the items are screened automatically and by an expert.

# Declarations

**CRediT.** Z. Geradts is the sole author and was responsible for conceptualisation, methodology, investigation and writing.

**Funding.** This work was funded from the internal research budget of the Netherlands Forensic Institute.

**Competing interests.** None.

**Ethics.** No new data of real persons were collected. FairFace, BFW, RFW and SCface were used under their licences and processed locally, and only group averages are reported. The synthetic faces were checked against real faces for memorisation.

**Data and code availability.** The code is available at https://github.com/geradts1/facebias (MIT licence), together with Supplementary Tables S1–S10 (supplement/Supplementary_Tables_S1-S10.xlsx). The synthetic image set will be released with the journal version. FairFace, BFW, RFW and SCface can be obtained from their owners under their own licences.

**Use of AI.** Generative image models were used to create the synthetic data set. An AI assistant (Claude) helped with programming and with drafting the text. I checked all results and the full text and take full responsibility for them.

# Supplementary material

The supplementary tables are provided as one spreadsheet in the code repository. Table S1 describes the synthetic identities, with the number of persons and the distribution of attributes per ancestry group. Table S2 gives the failure-to-enrol rates per group and condition for SCRFD and YuNet. Tables S3 and S4 contain, for every recogniser, group and condition class, the FMR and FNMR with 95% confidence intervals and the LR shift with the rates of misleading evidence for global and per-group calibration. Table S5 gives the results of the mixed model and of the quality-aware calibration, and Table S6 those for twins and look-alikes. Table S7 contains the sensitivity analyses for cohort normalisation (top-k of 50, 100 and 200 and a cohort of 600 or 1,200) and for the duplicate cut-off on FairFace. Table S8 gives the results on real faces (FairFace, BFW and RFW) per group, Table S9 the full comparison with the second generator, and Table S10 the near-duplicate and homogeneity check of the generated identities.